\documentclass{article}
\pdfoutput=1

\PassOptionsToPackage{numbers}{natbib}
\usepackage[final]{neurips_ai4d3_2023}

\usepackage{pgfplots}
\pgfplotsset{compat=1.18}
\usepackage{pgfplotstable}
\usepackage{subcaption}
\usepackage{amsmath}
\usepackage{amsthm}
\usepackage{marginnote}
\usepackage{multicol}
\usepackage{xcolor}
\usepackage{wrapfig}
\usepackage{algorithm}
\usepackage{algpseudocode}

\definecolor{col0}{rgb}{0.4,0.2,0.2}
\definecolor{col1}{rgb}{0.6,0.4,0.2}
\definecolor{col2}{rgb}{0.8,0.6,0.2}
\definecolor{col3}{rgb}{0.0,0.5,1.0}
\definecolor{lightgrey}{rgb}{0.8,0.8,0.8}
\definecolor{col4}{rgb}{0.66,0.5,0.83}
\definecolor{col5}{rgb}{0.83,0.5,0.66}
\definecolor{col6}{rgb}{1.0,0.5,0.0}

\newtheorem{definition}{Definition}

\usepackage[utf8]{inputenc} 
\usepackage[T1]{fontenc}    
\usepackage{hyperref}       
\usepackage{url}            
\usepackage{booktabs}       
\usepackage{amsfonts}       
\usepackage{nicefrac}       
\usepackage{microtype}      
\usepackage{xcolor}         

\title{Automating reward function configuration for drug design}

\author{%
    Marius Urbonas$^1$ \quad Temitope Ajileye$^2$ \quad Paul Gainer$^2$ \quad Douglas Pires$^2$ \\
    $^1$University of Oxford \quad $^2$Exscientia \\
    \texttt{marius.urbonas@stx.ox.ac.uk}  \\
    \texttt{\{tajileye,pgainer,dpires\}@exscientia.co.uk}
}

\begin{document}

\newcommand{\component}{\mathrm{c}}
\newcommand{\components}{\mathcal{C}}
\newcommand{\normaliser}{\phi}
\newcommand{\assay}{\mathrm{\alpha}}
\newcommand{\assays}{\mathcal{A}}

\newcommand{\params}{\theta}
\newcommand{\reward}{r}
\newcommand{\rewardfunc}{R}
\newcommand{\mols}{\mathcal{M}}
\newcommand{\project}{\mathsf{Project}} 

\newcommand{\mol}{m}
\newcommand{\target}{\mathfrak{T}}
\newcommand{\evalfunc}{E}

\newcommand{\dom}{\mathrm{dom}}
\newcommand{\pairs}[1]{\mathcal{D}_{\hat{t}}(#1)}

\newcommand{\rewardneural}{\hat{\reward}_{\params}}
\newcommand{\loss}{\mathcal{L}}
\newcommand{\weights}{\mathbf{w}}

\newcommand{\gmdtool}{\mathcal{G}}

\newcommand{\prob}{P}

\newcommand{\nats}{\mathbb{N}}
\newcommand{\reals}{\mathbb{R}}

\makeatletter
\newcommand\resetstackedplots{
\makeatletter
\pgfplots@stacked@isfirstplottrue
\makeatother
\addplot [forget plot,draw=none] coordinates{};
}
\makeatother

\newcommand{\errorband}[5][]{ 
\pgfplotstableread{#2}\datatable
    \addplot [draw=none, stack plots=y, forget plot] table [
        x={#3},
        y expr=\thisrow{#4}-\thisrow{#5}
    ] {\datatable};

    \addplot [draw=none, forget plot, fill=gray!40, stack plots=y, area legend, #1] table [
        x={#3},
        y expr=2*\thisrow{#5}
    ] {\datatable} \closedcycle;
    

    \resetstackedplots
}

\newcommand{\plotheight}{0.6\textwidth}
\newcommand{\figwidth}{0.475\textwidth}
\newcommand{\baseline}[1]{
    \addplot [domain=0:20, thick, black] {#1};
}

\usetikzlibrary{patterns}.

\maketitle

\begin{abstract}
Designing reward functions that can guide generative molecular design (GMD) algorithms to desirable areas of chemical space is of critical importance in AI-driven drug discovery. Traditionally, this has been a manual and error-prone task; the selection of appropriate computational methods to approximate biological assays is challenging and the normalisation and aggregation of computed values into a single score even more so, leading to potential reliance on trial-and-error approaches~\cite{sundin2022human}. We propose a novel approach for automated reward configuration that relies solely on experimental data, mitigating the challenges of manual reward adjustment on drug discovery projects. Our method achieves this by constructing a ranking over experimental data based on Pareto dominance over the multi-objective space, then training a neural network to approximate the reward function such that rankings determined by the predicted reward correlate with those determined by the Pareto dominance relation. We validate our method using two case studies. In the first study we simulate Design-Make-Test-Analyse (DMTA) cycles by alternating reward function updates and generative runs guided by that function. We show that the learned function adapts over time to yield compounds that score highly with respect to evaluation functions taken from the literature~\cite{brown2019guacamol}. In the second study we apply our algorithm to historical data from four real drug discovery projects. We show that our algorithm yields reward functions that outperform the predictive accuracy of human-defined functions, achieving an improvement of up to $0.4$ in Spearman's correlation against a ground truth evaluation function that encodes the target drug profile for that project. Our method provides an efficient data-driven way to configure reward functions for GMD, and serves as a strong baseline for future research into transformative approaches for the automation of drug discovery.
\end{abstract}

\section{Introduction}

Drug discovery is intrinsically a multi-objective optimisation problem~\cite{d2012multi}. Generative molecular design (GMD) algorithms explore the extensive space of drug-like molecules searching for compounds with desirable property profiles, simultaneously maximising or minimising multiple objectives~\cite{luukkonen2023artificial}. Many GMD algorithms are designed to generate molecules that maximise a given reward function~\cite{blaschke2020reinvent,gottipati2020learning,zhou2019optimization}.
Reward functions are a way for designers to communicate their goals to their GMD tools, and directly impact how those tools explore chemical space. Therefore, defining reward functions that can lead to the generation of compounds that meet the goals of the project is of critical importance in drug discovery. However, translating project goals into appropriate reward functions is a challenging task. Decisions must be made on which combinations of computable \textit{in silico} methods are appropriate to function as good proxies for the results of comparatively slow and expensive biological assays, and on how to normalise these computed values and combine them into a single score.

When reward functions consist of many components it becomes difficult to quantify the relationships between parameters. Humans in particular struggle at this task, just as they would struggle to quantify the relationships between the features in a high dimensional regression task. Sundin \textit{et al.}~\cite{sundin2022human}
observe that the result of this can often be that drug designers may resort to trial and error to explore the parameter space. To alleviate this they introduce a method to integrate human feedback into the optimisation of reward functions employed to guide the REINVENT tool~\cite{blaschke2020reinvent} using active learning. The main challenge addressed is the creation of a reward function that aligns with a human chemist's optimisation goals, and the proposed method allows the reward function to be learned directly from user feedback, eliminating the need for manual tuning through trial and error. While this method is shown to be effective for molecular generation, it relies on human expertise to assign binary preferences to generated compounds, and it has been demonstrated that there is low consensus between chemists when prioritising molecules~\cite{gomez2018decision,kutchukian2012inside}.

To eliminate the dependence on biased human feedback we propose a novel approach that relies solely on the results of biological assays to guide the learning of a suitable reward function. To achieve this we build on advancements in the field of Inverse Reinforcement Learning (IRL), that we use as the backbone of our method. IRL is the problem of learning a reward function by observing a set of expert demonstrations, where the optimality of any learned function depends on the quality of the demonstrations~\cite{ng2000algorithms}. Agyemang \textit{et al}.~\cite{agyemang2021deep} use entropy maximisation to learn a reward model to guide a Reinforcement Learning agent to optimise single-objective tasks for molecular generation, including the octanol-water partition coefficient (logP) and binding to the DRD2 protein. They show that their approach is effective when high-quality molecular data exhibiting the objective is readily available. However, when optimising for multiple objectives it is difficult to obtain a set of high-quality demonstrations. For real drug discovery projects, available molecular data is usually scarce or nonexistent, and expensive to augment as the project progresses due to the high cost of conducting biological assays. This is particularly challenging during the early stages of a project, where only a handful of noisy observations are available.

To address this problem, we adopt the approach of Brown \textit{et al.}~\cite{brown2019extrapolating} where a reward function that yields better-than-demonstrator performance is obtained by learning to justify preferential rankings over demonstrations, rather than learning to imitate them. There are two main approaches to construct preference rankings over molecules for multi-parameter optimisation tasks~\cite{luukkonen2023artificial}: aggregation methods and Pareto-based methods. The first approach combines multiple objective functions into a single score that can be used to rank the molecules. However, the manual specification of the scalarisation function faces the same challenges as that of manually specifying a reward function, and encodes the bias of human chemists when exploring trade-offs between objectives.

The main contribution of this paper is a principled Pareto-based method to learn reward functions that relies solely on experimental assay data. We achieve this by defining partial rankings using pairwise Pareto dominance relationships, where one molecule is preferred to (\emph{dominates}) another if it as least as good for every objective, and better for at least one objective. We then use these partial rankings to train the reward function as a preference predictor over molecules. We demonstrate empirically that our method can learn reward functions that yield rankings that outperform human-defined functions over data from real drug discovery projects. Furthermore, we apply our method to synthetic goals taken from a well established benchmark for GMD, and show that we can learn reward functions that can guide a method taken from that benchmark to generate compounds that score as highly as those generated using the target reward function.

In Section~\ref{sec:method} we describe our proposed method, which we then evaluate using data from public benchmarks and real drug discovery projects in Section~\ref{sec:experiments}. We conclude the paper with a discussion of our results and an outline of future directions for work in Section~\ref{sec:discussion}.
\section{Methods}
\label{sec:method}



Let $\mols = \{\mol_1, \mol_2,\cdots,\}$ be a set of molecules, and let $\assay_1,\cdots, \assay_K$ be real valued evaluation functions over the molecular space: these functions map each molecule into $\reals^K$. We will often refer to these functions as assays and think of molecules
as points in that multi-dimensional space. Experimentally characterising each molecule designed by a discovery team is infeasible; the decision of which subset to evaluate is guided by a reward function $\reward$, ideally inexpensive to compute, that gives higher scores to molecules that are closer to $\target$ in evaluation space.

Our method consists of two main steps. In the first step we construct rankings in $\reals^K$ based on Pareto dominance: this encodes the notion that, lacking any overriding directives, $\mol_1$ can only be said to be preferred over $\mol_2$ if it is closer to $\target$ on all assays. In the second step we train a neural network to generate a reward function such that the rankings determined by the function correlate with those constructed in the previous step. The neural network is given access to the components that humans would usually include in their reward functions.

\begin{wrapfigure}{r}{0.5\textwidth} 
    \begin{center}
        \begin{tikzpicture}
            \begin{axis}
                [
                    xmin=0,
                    xmax=1,
                    ymin=0,
                    ymax=1,
                    axis lines=middle,
                    enlargelimits=true,
                    xmajorticks=false,
                    ymajorticks=false,
                    xlabel={$\assay_1$},
                    ylabel={$\assay_2$},
                    x label style={at={(axis description cs:0.5, 0.05)},anchor=north},
                    y label style={at={(axis description cs:0.05, .5)},rotate=90,anchor=south},
                    width=0.48\textwidth,
                    height=0.48\textwidth,
                ]
                \addplot[
                    only marks,
                    mark=*,
                    mark size=2pt,
                    fill=lightgrey,
                ]
                table{data/pareto_points.dat};
                \addplot[
                    only marks,
                    mark=*,
                    mark size=2pt,
                    fill=col2,
                    draw=col1,
                ]
                table{
                    x y
                    0.3 0.75
                };
                \addplot[
                    mesh,
                    thick,
                    col1,
                    dashed,
                ]
                table{
                    x y
                    0.0 0.75
                    0.3 0.75
                    0.3 0.0
                };
                \addplot[
                    no marks,
                    fill,
                    col1,
                    fill opacity=0.1,
                ]
                table{
                    x y
                    0.0 0.75
                    0.3 0.75
                    0.3 0.0
                    0.0 0.0
                }
                \closedcycle;
                \node at (36, 79) {$\mol_1$};
                
                \addplot[
                    only marks,
                    mark=*,
                    mark size=2pt,
                    fill=col2,
                    draw=col1,
                ]
                table{
                    x y
                    0.42 0.45
                };
                \addplot[
                    mesh,
                    thick,
                    col1,
                    dashed,
                ]
                table{
                    x y
                    0.0 0.45
                    0.42 0.45
                    0.42 0.0
                };
                \addplot[
                    no marks,
                    fill,
                    col1,
                    fill opacity=0.1,
                ]
                table{
                    x y
                    0.0 0.45
                    0.42 0.45
                    0.42 0.0
                    0.0 0.0
                }
                \closedcycle;
                \node at (48, 50) {$\mol_2$};
                
                \addplot[
                    only marks,
                    mark=*,
                    mark size=2pt,
                    fill=col2,
                    draw=col1,
                ]
                table{
                    x y
                    0.82 0.2
                };
                \addplot[
                    mesh,
                    thick,
                    col1,
                    dashed,
                ]
                table{
                    x y
                    0.0 0.2
                    0.82 0.2
                    0.82 0.0
                };
                \addplot[
                    no marks,
                    fill,
                    col1,
                    fill opacity=0.1,
                ]
                table{
                    x y
                    0.0 0.2
                    0.82 0.2
                    0.82 0.0
                    0.0 0.0
                }
                \closedcycle;
                \node at (88, 25) {$\mol_3$};
            \end{axis}
        \end{tikzpicture}
        \caption{Example of Pareto dominance relationships: for each molecule $\mol_i$ the corresponding shaded region shows the set $\dom(\mol_i)$ of molecules dominated by $\mol_i$ in the multi-objective space defined by the results of two biological assays $\assay_1$ and $\assay_2$.}
        \label{fig:pareto}
    \end{center}
\end{wrapfigure}
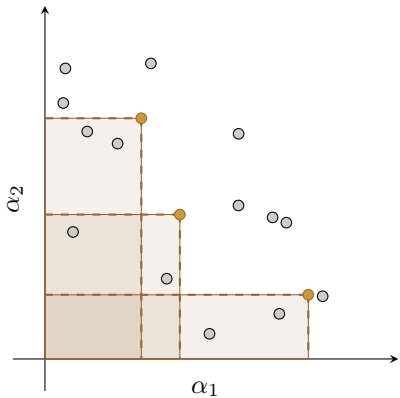

\subsection{Pareto Dominance Ranking}
Traditionally, in order to establish a preference relationship between two molecules, designers encode an opinion on the scaling functions and relative weights of each assay~\citep{luukkonen2023artificial}, so that an average of the results can be used to determine preferences. Our method only requires the selection of relevant evaluation functions, rather than the specification of those parameters, and as a consequence is less biased.

The evaluation-based ranking over $\mols$ is constructed by determining Pareto dominance relationships over the multi-objective space defined by the results of $K$ biological assays with respect to some point $\target{=}(t_1{,}t_2{,}\ldots{,}t_k)$ that corresponds to a drug candidate criteria, where each $t_i \in \reals$ is the target value for assay $\assay_i$, as illustrated in Figure~\ref{fig:pareto}.
\begin{definition}
Given two molecules $\mol_1$ and $\mol_2$, and a drug candidate criteria $\target = (t_1, t_2, \ldots, t_K)$, we say that $\mol_1$ \emph{dominates} $\mol_2$ if $| \assay_i(\mol_1)-t_i| < |\assay_i(\mol_2)-t_i|$ for all $1 \le i \le K$. We denote this relationship by $\mol_1 \succ \mol_2$.
\end{definition}
The set $\pairs{\mols}$ of all domination pairs of molecules can then be constructed as:
\begin{equation}
    \pairs{\mols} = \bigcup_{\mol_1 \in \mols} \{(\mol_1, \mol_2) \mid \mol_2 \in \dom(\mol_1)\},
\end{equation}
where $\dom(\mol)$ denotes the set of molecules in $\mols$ that are dominated by $\mol$.

\subsection{The Reward Network}
The learnable reward function is a composition of \emph{components} $\{\component_i\}_{i = 1, \ldots, N}$, where each component is a physics-based, machine-learned, or otherwise-computable function mapping compounds to scores in $\reals$. Components are unbounded, therefore for each $\component_i$ there is a corresponding normalising function $\phi_i: \reals \to [0, 1]$. To ensure interpretability and reliability of any learned function, we fix the shape of $\rewardneural$ to be a linear combination over components:
\begin{align}
\label{eq:reward}
\rewardneural(\mol) = \weights \cdot
    \left[
        \begin{array}{c}
             \normaliser_1(\component_1(\mol)) \\
             \normaliser_2(\component_2(\mol)) \\
             \ldots \\
             \normaliser_N(\component_N(\mol)) 
        \end{array}
    \right]
\end{align}

We follow~\citep{sundin2022human} and use two different normalising functions: a sigmoid function to bound a maximisation property and a Gaussian activation to bound a property that is expected to have an optimal range. In both cases the learning framework estimates the parameters of these functions; slope and midpoint for the sigmoid, and mean and variance for the Gaussian activation. 

To train $\rewardneural$ on preference pairs we follow the approach adopted in~\citep{brown2019extrapolating}, and model the probability of a molecule being preferred as depending exponentially on the value of the reward using a softmax-normalised distribution:
\begin{align}
    \prob(\mol_1 \succ \mol_2) & \approx
        \frac
        {\exp \rewardneural(\mol_1)}
        {\exp \rewardneural(\mol_1) + \exp \rewardneural(\mol_2)}.
\end{align}

The cross-entropy between predictions from $\rewardneural$ and the Pareto dominance relation pairs $\pairs{\mols}$ is then minimised using the following loss function that has been demonstrated to be effective for training models from preferences~\citep{brown2019extrapolating,christiano2017deep,ibarz2018reward}:
\begin{align}
\label{eq:loss}
    \loss(\params)
        & = -\sum_{\mol_1 \succ \mol_2} \log  \prob(\mol_1 \succ \mol_2) \\
        & = -\sum_{\mol_1 \succ \mol_2} \log
        \frac
            {\exp \rewardneural(\mol_1)}
            {\exp \rewardneural(\mol_1) + \exp \rewardneural(\mol_2)}.
\end{align}


\section{Experiments and Results}
\label{sec:experiments}

We conduct two experiments to evaluate our method. In the first we investigate the effectiveness of our learned reward functions for guiding a generative method to desirable regions of molecular space. To achieve this we simulate DMTA cycles by defining synthetic proxies for project goals using reward functions taken from a publicly available benchmark for GMD. For the second experiment we apply our method to data taken from four real drug discovery projects, and evaluate its performance when compared to human-defined reward functions taken from those projects. For our method to be successful in this experiment, we expect rankings obtained using our method to have a higher correlation with project goals than rankings obtained using human defined reward functions.

\subsection{Experiments on Simulated Design Cycles}

We first evaluate the effectiveness of our method when used to guide a GMD algorithm towards desirable regions of molecular space. To simulate the cycles of a drug discovery project we extract a set of reward functions from the GuacaMol~\cite{brown2019guacamol} benchmark that serve as proxies for drug discovery project goals. We use these functions as evaluation functions for determining of true rankings. We identify $6$ reward functions with which to evaluate our method from the goal-directed section of that benchmark. Of the $20$ available functions, $9$ are single parameter objectives, and hence out of scope for what we are aiming to show, $2$ have primarily binary scoring components, and hence are not representative of the continuous scores that result from biological assays, and $2$ consist of simple maximisation objectives. Finally, we omit the remaining \emph{Ranolazine MPO} function as the components for that function reported in the paper differ from those available in the corresponding data repository.

\begin{algorithm}
\caption{Experiment procedure for simulated DMTA cycles. }\label{alg:guacamol}
\begin{algorithmic}
\Require$\evalfunc$ an evaluation function
\Require $\target$ the drug candidate criteria derived from $E$
\Require $\gmdtool$ a GMD tool
\Require $T$ number of repetitions
\For{$t = 1 \dots T$}
\State $\mols \gets$ a random sample of 20 molecules from ChEMBL
\For{$i = 1 \dots 20$}
    \State compute $\pairs{\mols}$ and use to train $\rewardneural$
    \State $\mols_i \gets$ molecules generated by $\gmdtool$ using $\rewardneural$ and seed molecules $\mols$
    \State $\mols_i^+ \gets$ top 10 molecules in $\mols_i$ or a diverse set  of 10
    \State $\mols \gets \mols \cup \mols_i^+$
    \State compute the mean and max values obtained by evaluating $\evalfunc$ over $\mols_i$
\EndFor
\EndFor
\end{algorithmic}
\end{algorithm}

For each reward function from the benchmark we construct a drug candidate criteria $\target$ by reading the values from the reward function components. We set the components of the learnable reward function to be those of the target function. However, we initialise the parameters for aggregation and scaling for the learnable function by sampling from a unit Gaussian. Algorithm~\ref{alg:guacamol} shows the procedure used to simulate DMTA cycles and evaluate our approach. To simulate the sub-optimal molecules available at the start of a drug discovery project we initially sample 20 molecules uniformly at random from the subset of ChEMBL~\cite{gaulton2012chembl} molecules used for that task in \cite{brown2019guacamol}. We then simulate $20$ cycles by repeatedly constructing preferences over the existing set of evaluated molecules. For each cycle we fit a reward function using Pareto-dominance pairs, use a GMD method to optimise that function and generate new molecules, and evaluate the top $10$ scoring molecules. We repeat the whole experiment three times, and aggregate the results.

We use the highest performing GMD method (\emph{Graph GA}) from~\cite{brown2019guacamol} to evaluate our approach. Graph GA is a genetic algorithm that follows the implementation of Jensen~\cite{jensen2019graph}. We configure it with a population size of $200$, an offspring size of $200$, and mutation rate of $0.01$, and run for $100$ generations at each iteration. We run two versions of this experiment. In the first, we select the $10$ top scoring molecules after each iteration to use for subsequent iterations. In the second, we cluster all generated molecules at each iteration by their ECFP6 fingerprint~\cite{rogers2010extended} using the K-means clustering algorithm and selecting the molecules closest to the cluster centres to encourage diversity.

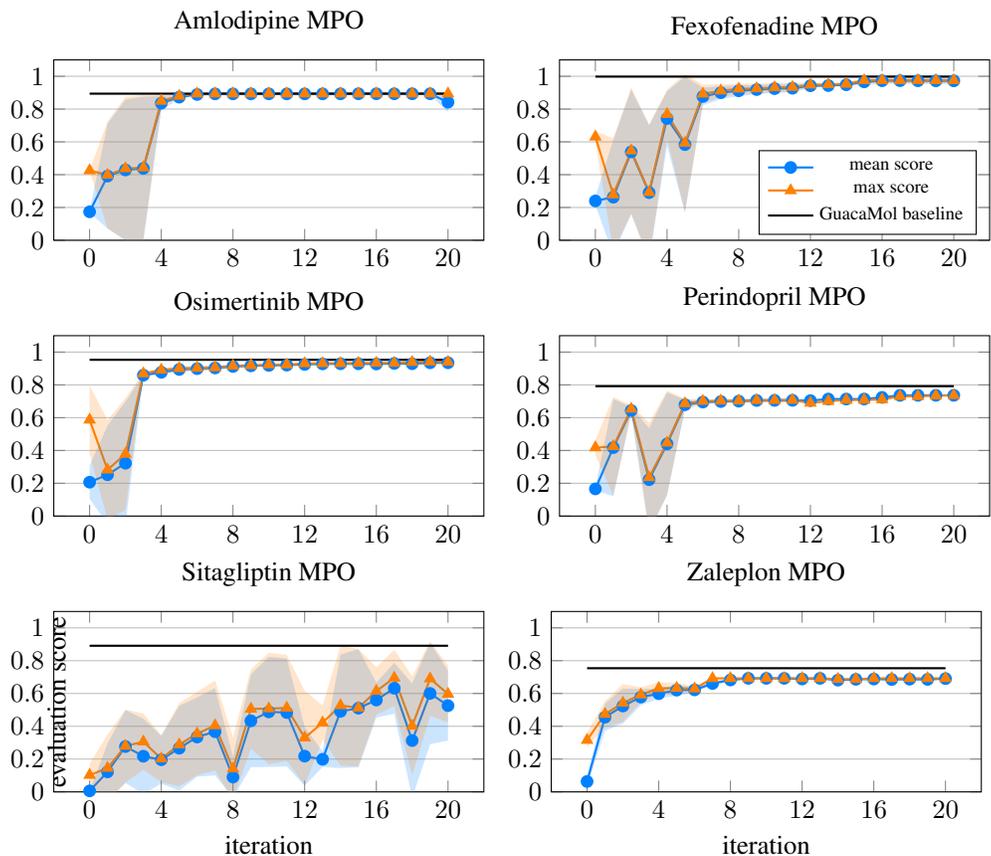
\begin{figure}[h]
    \captionsetup[subfigure]{labelformat=empty}
    \begin{subfigure}{\figwidth}
        \centering
        \begin{tikzpicture}
            \begin{axis}[
                ymajorgrids=true,
                ylabel={\phantom{label}},
                xtick={0,4,...,20},
                ytick={0,0.2,...,1},
                ymin=0,
                ymax=1.1,
                legend pos=south east,
                title={Amlodipine MPO},
                width=1.1\textwidth,
                height=\plotheight,
                y label style={at={(0.05,0.5)}},
            ]
                \addplot [col3,mark=*,thick] table [x index=0, y index=1]{data/amlodipine_mean.dat};
                \addplot [col6,mark=triangle*,thick] table [x index=0, y index=1] {data/amlodipine_max.dat};
                \errorband[col3, opacity=0.2]{data/amlodipine_mean.dat}{x}{y}{error}
                \errorband[col6, opacity=0.2]{data/amlodipine_max.dat}{x}{y}{error}
                \baseline{0.894}
            \end{axis}
        \end{tikzpicture}
    \end{subfigure}
    \begin{subfigure}{\figwidth}
        \centering
        \begin{tikzpicture}
            \begin{axis}[
                ymajorgrids=true,
                ylabel={\phantom{label}},
                xtick={0,4,...,20},
                ytick={0,0.2,...,1},
                ymin=0,
                ymax=1.1,
                legend pos=south east,
                title={Fexofenadine MPO},
                width=1.1\textwidth,
                height=\plotheight,
                legend style={font=\scriptsize},
                y label style={at={(0.05,0.5)}},
            ]
                \addplot [col3,mark=*,thick] table [x index=0, y index=1]{data/fexofenadine_mean.dat};
                \addlegendentry{mean score}
                \addplot [col6,mark=triangle*,thick] table [x index=0, y index=1] {data/fexofenadine_max.dat};
                \addlegendentry{max score}
                \errorband[col3, opacity=0.2]{data/fexofenadine_mean.dat}{x}{y}{error}
                \errorband[col6, opacity=0.2]{data/fexofenadine_max.dat}{x}{y}{error}
                \baseline{0.998}
                \addlegendentry{GuacaMol baseline}
            \end{axis}
        \end{tikzpicture}
    \end{subfigure}
    \begin{subfigure}{\figwidth}
        \centering
        \begin{tikzpicture}
            \begin{axis}[
                ymajorgrids=true,
                ylabel={\phantom{label}},
                xtick={0,4,...,20},
                ytick={0,0.2,...,1},
                ymin=0,
                ymax=1.1,
                legend pos=south east,
                title={Osimertinib MPO},
                width=1.1\textwidth,
                height=\plotheight,
                y label style={at={(0.05,0.5)}},
            ]
                \addplot [col3,mark=*,thick] table [x index=0, y index=1]{data/osimertinib_mean.dat};
                \addplot [col6,mark=triangle*,thick] table [x index=0, y index=1] {data/osimertinib_max.dat};
                \errorband[col3, opacity=0.2]{data/osimertinib_mean.dat}{x}{y}{error}
                \errorband[col6, opacity=0.2]{data/osimertinib_max.dat}{x}{y}{error}
                \baseline{0.953}
            \end{axis}
        \end{tikzpicture}     
    \end{subfigure}
    \begin{subfigure}{\figwidth}
        \centering
        \begin{tikzpicture}
            \begin{axis}[
                ymajorgrids=true,
                ylabel={\phantom{label}},
                xtick={0,4,...,20},
                ytick={0,0.2,...,1},
                ymin=0,
                ymax=1.1,
                legend pos=south east,
                title={Perindopril MPO},
                width=1.1\textwidth,
                height=\plotheight,
                y label style={at={(0.05,0.5)}},
            ]
                \addplot [col3,mark=*,thick] table [x index=0, y index=1]{data/perindopril_mean.dat};
                \addplot [col6,mark=triangle*,thick] table [x index=0, y index=1] {data/perindopril_max.dat};
                \errorband[col3, opacity=0.2]{data/perindopril_mean.dat}{x}{y}{error}
                \errorband[col6, opacity=0.2]{data/perindopril_max.dat}{x}{y}{error}
                \baseline{0.792}
            \end{axis}
        \end{tikzpicture}       
    \end{subfigure}
    \begin{subfigure}{\figwidth}
        \centering
        \begin{tikzpicture}
            \begin{axis}[
                ymajorgrids=true,
                xlabel={iteration},
                ylabel={evaluation score},
                xtick={0,4,...,20},
                ytick={0,0.2,...,1},
                ymin=0,
                ymax=1.1,
                legend pos=south east,
                title={Sitagliptin MPO},
                width=1.1\textwidth,
                height=\plotheight,
                y label style={at={(0.05,0.5)}},
            ]
                \addplot [col3,mark=*,thick] table [x index=0, y index=1]{data/sitagliptin_mean.dat};
                \addplot [col6,mark=triangle*,thick] table [x index=0, y index=1] {data/sitagliptin_max.dat};
                \errorband[col3, opacity=0.2]{data/sitagliptin_mean.dat}{x}{y}{error}
                \errorband[col6, opacity=0.2]{data/sitagliptin_max.dat}{x}{y}{error}
                \baseline{0.891}
            \end{axis}
        \end{tikzpicture}            
    \end{subfigure}
    \centering
    \begin{subfigure}{\figwidth}
        \begin{tikzpicture}
            \begin{axis}[
                ymajorgrids=true,
                xlabel={iteration},
                ylabel={\phantom{label}},
                xtick={0,4,...,20},
                ytick={0,0.2,...,1},
                ymin=0,
                ymax=1.1,
                legend pos=south east,
                title={Zaleplon MPO},
                width=1.1\textwidth,
                height=\plotheight,
                y label style={at={(0.05,0.5)}},
            ]
                \addplot [col3,mark=*,thick] table [x index=0, y index=1]{data/zaleplon_mean.dat};
                \addplot [col6,mark=triangle*,thick] table [x index=0, y index=1] {data/zaleplon_max.dat};
                \errorband[col3, opacity=0.2]{data/zaleplon_mean.dat}{x}{y}{error}
                \errorband[col6, opacity=0.2]{data/zaleplon_max.dat}{x}{y}{error}
                \baseline{0.754}
            \end{axis}
        \end{tikzpicture}            
    \end{subfigure}
    \caption{Mean and maximum evaluation scores for molecules generated at each iteration of Graph GA using the top scoring selection method. The shaded regions show variance across repeated runs.}
    \label{fig:highest}
\end{figure}

\begin{figure}[h]
    \captionsetup[subfigure]{labelformat=empty}
    \begin{subfigure}{\figwidth}
        \centering
        \begin{tikzpicture}
            \begin{axis}[
                ymajorgrids=true,
                ylabel={\phantom{label}},
                xtick={0,4,...,20},
                ytick={0,0.2,...,1},
                ymin=0,
                ymax=1.1,
                legend pos=south east,
                title={Amlodipine MPO},
                width=1.1\textwidth,
                height=\plotheight,
                y label style={at={(0.05,0.5)}},
            ]
                \addplot [col3,mark=*,thick] table [x index=0, y index=1]{data/amlodipine_mean_cluster.dat};
                \addplot [col6,mark=triangle*,thick] table [x index=0, y index=1] {data/amlodipine_max_cluster.dat};
                \errorband[col3, opacity=0.2]{data/amlodipine_mean_cluster.dat}{x}{y}{error}
                \errorband[col6, opacity=0.2]{data/amlodipine_max_cluster.dat}{x}{y}{error}
                \baseline{0.894}
            \end{axis}
        \end{tikzpicture}
    \end{subfigure}
    \begin{subfigure}{\figwidth}
        \centering
        \begin{tikzpicture}
            \begin{axis}[
                ymajorgrids=true,
                ylabel={\phantom{label}},
                xtick={0,4,...,20},
                ytick={0,0.2,...,1},
                ymin=0,
                ymax=1.1,
                legend pos=south east,
                title={Fexofenadine MPO},
                width=1.1\textwidth,
                height=\plotheight,
                legend style={font=\scriptsize},
                y label style={at={(0.05,0.5)}},
            ]
                \addplot [col3,mark=*,thick] table [x index=0, y index=1]{data/fexofenadine_mean_cluster.dat};
                \addlegendentry{mean evaluation}
                \addplot [col6,mark=triangle*,thick] table [x index=0, y index=1] {data/fexofenadine_max_cluster.dat};
                \addlegendentry{max evaluation}
                \errorband[col3, opacity=0.2]{data/fexofenadine_mean_cluster.dat}{x}{y}{error}
                \errorband[col6, opacity=0.2]{data/fexofenadine_max_cluster.dat}{x}{y}{error}
                \baseline{0.998}
                \addlegendentry{GuacaMol baseline}
            \end{axis}
        \end{tikzpicture}
    \end{subfigure}
    \begin{subfigure}{\figwidth}
        \centering
        \begin{tikzpicture}
            \begin{axis}[
                ymajorgrids=true,
                ylabel={\phantom{label}},
                xtick={0,4,...,20},
                ytick={0,0.2,...,1},
                ymin=0,
                ymax=1.1,
                legend pos=south east,
                title={Osimertinib MPO},
                width=1.1\textwidth,
                height=\plotheight,
                y label style={at={(0.05,0.5)}},
            ]
                \addplot [col3,mark=*,thick] table [x index=0, y index=1]{data/osimertinib_mean_cluster.dat};
                \addplot [col6,mark=triangle*,thick] table [x index=0, y index=1] {data/osimertinib_max_cluster.dat};
                \errorband[col3, opacity=0.2]{data/osimertinib_mean_cluster.dat}{x}{y}{error}
                \errorband[col6, opacity=0.2]{data/osimertinib_max_cluster.dat}{x}{y}{error}
                \baseline{0.953}
            \end{axis}
        \end{tikzpicture}     
    \end{subfigure}
    \begin{subfigure}{\figwidth}
        \centering
        \begin{tikzpicture}
            \begin{axis}[
                ymajorgrids=true,
                ylabel={\phantom{label}},
                xtick={0,4,...,20},
                ytick={0,0.2,...,1},
                ymin=0,
                ymax=1.1,
                legend pos=south east,
                title={Perindopril MPO},
                width=1.1\textwidth,
                height=\plotheight,
                y label style={at={(0.05,0.5)}},
            ]
                \addplot [col3,mark=*,thick] table [x index=0, y index=1]{data/perindopril_mean_cluster.dat};
                \addplot [col6,mark=triangle*,thick] table [x index=0, y index=1] {data/perindopril_max_cluster.dat};
                \errorband[col3, opacity=0.2]{data/perindopril_mean_cluster.dat}{x}{y}{error}
                \errorband[col6, opacity=0.2]{data/perindopril_max_cluster.dat}{x}{y}{error}
                \baseline{0.792}
            \end{axis}
        \end{tikzpicture}       
    \end{subfigure}
    \begin{subfigure}{\figwidth}
        \centering
        \begin{tikzpicture}
            \begin{axis}[
                ymajorgrids=true,
                xlabel={iteration},
                ylabel={evaluation score},
                xtick={0,4,...,20},
                ytick={0,0.2,...,1},
                ymin=0,
                ymax=1.1,
                legend pos=south east,
                title={Sitagliptin MPO},
                width=1.1\textwidth,
                height=\plotheight,
                y label style={at={(0.05,0.5)}},
            ]
                \addplot [col3,mark=*,thick] table [x index=0, y index=1]{data/sitagliptin_mean_cluster.dat};
                \addplot [col6,mark=triangle*,thick] table [x index=0, y index=1] {data/sitagliptin_max_cluster.dat};
                \errorband[col3, opacity=0.2]{data/sitagliptin_mean_cluster.dat}{x}{y}{error}
                \errorband[col6, opacity=0.2]{data/sitagliptin_max_cluster.dat}{x}{y}{error}
                \baseline{0.891}
            \end{axis}
        \end{tikzpicture}            
    \end{subfigure}
    \centering
    \begin{subfigure}{\figwidth}
        \begin{tikzpicture}
            \begin{axis}[
                ymajorgrids=true,
                xlabel={iteration},
                ylabel={\phantom{label}},
                xtick={0,4,...,20},
                ytick={0,0.2,...,1},
                ymin=0,
                ymax=1.1,
                legend pos=south east,
                title={Zaleplon MPO},
                width=1.1\textwidth,
                height=\plotheight,
                y label style={at={(0.05,0.5)}},
            ]
                \addplot [col3,mark=*,thick] table [x index=0, y index=1]{data/zaleplon_mean_cluster.dat};
                \addplot [col6,mark=triangle*,thick] table [x index=0, y index=1] {data/zaleplon_max_cluster.dat};
                \errorband[col3, opacity=0.2]{data/zaleplon_mean_cluster.dat}{x}{y}{error}
                \errorband[col6, opacity=0.2]{data/zaleplon_max_cluster.dat}{x}{y}{error}
                \baseline{0.754}
            \end{axis}
        \end{tikzpicture}            
    \end{subfigure}
    \caption{Mean and maximum evaluation scores for molecules generated at each iteration of Graph GA using the K-means selection method. The shaded regions show variance across repeated runs.}
    \label{fig:kmeans}
\end{figure}
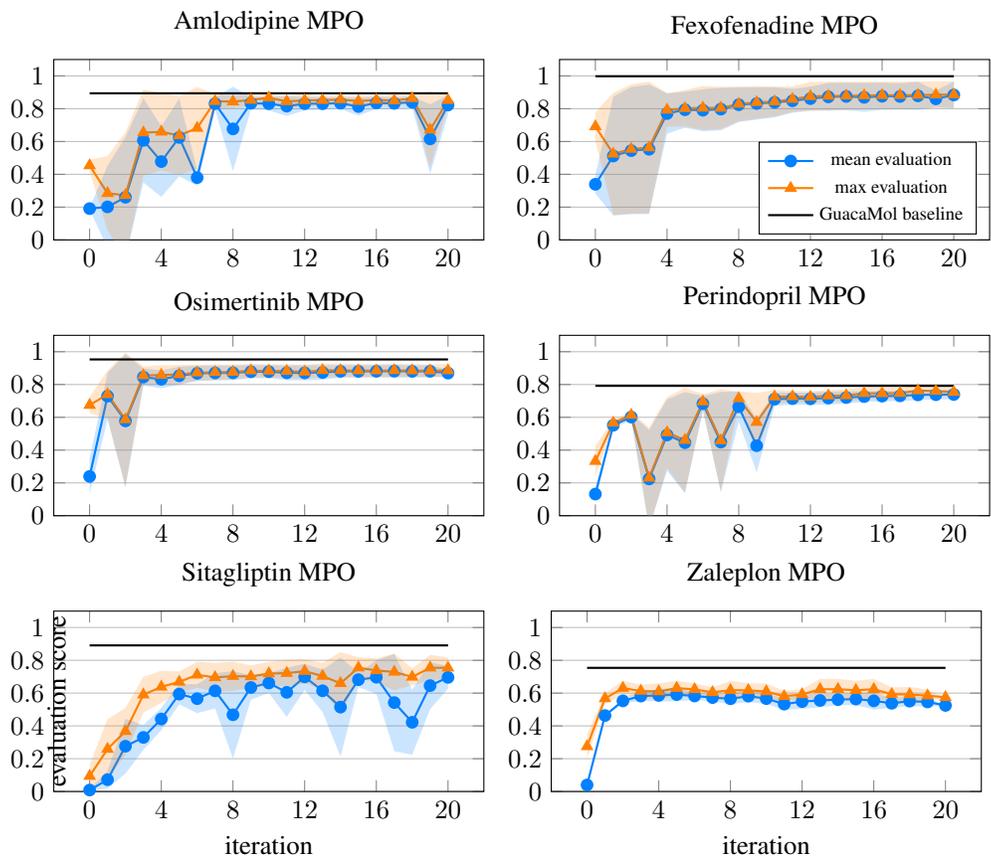

\paragraph{Results} The results for this experiment are plotted in Figures~\ref{fig:highest} and~\ref{fig:kmeans}. For each reward function that we seek to learn we plot a horizontal line indicating the performance of Graph GA in the benchmark, reported as the average over the mean scores of the top scoring molecule, the top $10$ scoring molecules and the top $100$ scoring molecules, when seeding generative runs with the \emph{highest scoring molecules} from the corresponding subset of ChEMBL. While that measure is not directly comparable to our mean scores, it represents the performance that we would expect if we used the evaluation function in place of the learned reward function.
For both selection strategies our method learns a reward function that guides Graph GA to molecules with a score comparable to those achieved in the benchmark within $10$ iterations, despite seeding our runs with a randomly initialised population from the dataset. When selecting top scoring compounds we achieve within $0.146$ of the baseline by iteration $10$ on average, and within $0.054$ by iteration $20$. For cluster selection we achieve within $0.138$ by iteration $10$, and within $0.108$ by cycle $20$. Performance was better across all functions when using top scoring compounds, with the exception of Sitagliptin. We postulate that this is due to this function having a larger number of Gaussian components, which are harder to optimise when molecules are selected by the same reward functions used to generate them.

\subsection{Experiments on Discovery Project Data}

We now evaluate our model on data taken from four active drug discovery projects (undisclosed targets):
$\project_1, \project_, \project_3$, and $\project_4$.
Each project has an associated target property profile  that specifies desirable ranges for assay results. If all assay results for a molecule satisfy this profile then the molecule is a potential drug candidate for that project. Therefore, we use the assays and ranges from that profile to define the drug candidate criteria $\target$ in the objective space.  We treat half-open intervals as minimization or maximization objectives by setting the target value to an arbitrary large or small float. For closed intervals we set the target value to the midpoint of the interval.

Each project is associated with a human-defined evaluation function $\evalfunc$ that aggregates normalised assay scores into a single score in the unit interval using the geometric mean. This ensures that molecules only score highly (close to $1$) when all of the assay component scores are close to the drug candidate criteria. Molecules with high scores are potential drug candidates. The evaluation function gives us another way to rank molecules based on experimental results. Although we do not train our method to maximise correlation with rankings determined by $\evalfunc$, we compute the Spearman's rank correlation coefficient between our learned rewards and $\evalfunc$ as a way to compare our method against an existing project-based measure of progress. 

For each project we obtain the most recent reward function configured by the discovery team on that project, and limit our method to only use components from that reward function. This limitation is imposed to assess the ability of the method to identify suitable weights and scaling parameters, and we defer an analysis of model performance when extending the set of available components to future work. We then take the set of molecules sent for synthesis for that project, along with their associated assay scores, and split it temporally across DMTA cycles. For each cycle $i$ we fit the model using rankings computed over the assay results for molecules up to cycle $i-1$, and perform inference on molecules from cycle $i$ to obtain a predicted ranking. Many of the components of these reward functions are predictive models that are trained on internal data. When a component of the reward function is a predictive model we retrain a restricted version of that model, limiting the training set by excluding data that was not available before the timestamp associated with cycle $i$.

Due to time limitations we do not use our restricted models for the evaluation of the human-defined reward function, and directly use the latest versions of those models from the projects. In addition, we do not include any of the foundational data from sources such as ChEMBL~\cite{gaulton2012chembl} that are often included in the training of those models. Therefore, in this experiment the human-defined reward function has access to predictive models with better performance than those that we use for training $\rewardneural$.

\begin{figure}
    \centering
    \begin{tikzpicture}
        \begin{axis}[
            xtick={5,7,...,29},
            ytick={-1.4,-1.2,...,1},
            extra y ticks=0,
            extra y tick style={grid=major, grid style={dashed,black}},
            xlabel=cycle,
            ylabel={$\Delta \rho$},
            legend style={draw=none},
            width=\textwidth,
            height=.32\textwidth,
            ymajorgrids=true,
            grid style={dotted,gray},
            legend pos=south east,
            xmin=4,
            xmax=30,
            legend style={font=\tiny},
        ]
            \addplot[
                col3,
                mark=*,
                error bars/.cd, 
                y fixed,
                y dir=both, 
                y explicit,
            ] table [x index=0, y index=14] {data/project_1.dat};
            \addlegendentry{$\project_1$} 
            \errorband[col3, opacity=0.2]{data/project_1.dat}{cycle}{spearmans_merit_mean_diff}{spearmans_merit_std};

            \addplot[
                col4,
                mark=triangle*,
                error bars/.cd, 
                y fixed,
                y dir=both, 
                y explicit,
            ] table [x index=0, y index=14] {data/project_2.dat};
            \addlegendentry{$\project_2$} 
            \errorband[col4, opacity=0.2]{data/project_2.dat}{cycle}{spearmans_merit_mean_diff}{spearmans_merit_std};

            \addplot[
                col5,
                mark=diamond*,
                error bars/.cd, 
                y fixed,
                y dir=both, 
                y explicit,
            ] table [x index=0, y index=14] {data/project_3.dat};
            \addlegendentry{$\project_3$} 
            \errorband[col5, opacity=0.2]{data/project_3.dat}{cycle}{spearmans_merit_mean_diff}{spearmans_merit_std}

            \addplot[
                col6,
                mark=square*,
                error bars/.cd, 
                y fixed,
                y dir=both, 
                y explicit,
            ] table [x index=0, y index=14] {data/project_4.dat};
            \addlegendentry{$\project_4$} 
            \errorband[col6, opacity=0.2]{data/project_4.dat}{cycle}{spearmans_merit_mean_diff}{spearmans_merit_std}
        \end{axis}
    \end{tikzpicture}
    \caption{Difference between the Spearman's correlation of $\evalfunc$ and learned reward functions rankings, and the Spearman's correlation of $\evalfunc$ and human-defined reward function rankings.}
    \label{figure:deltaspearmans}
\end{figure}
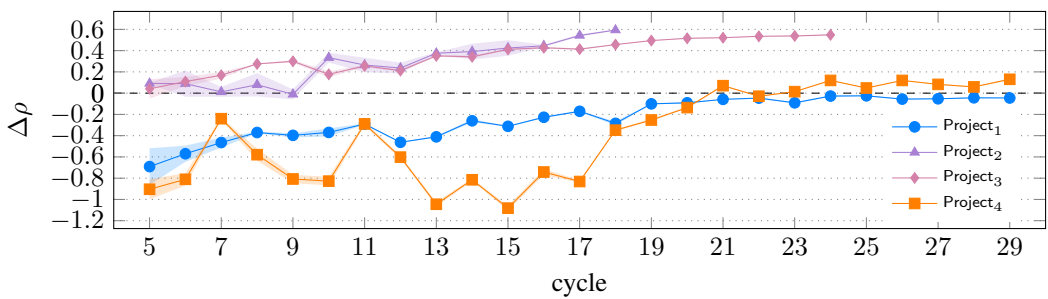

\paragraph{Results}
For each project, we compute the Spearman's rank correlation coefficient between our learned reward at each cycle and the evaluation function $\evalfunc$. We compute the same coefficient for the reference reward function and plot the difference as $\Delta \rho$ in Figure~\ref{figure:deltaspearmans}.
The results show that our method generates reward functions that meet or exceed the predictive power of the reference reward functions, despite the disadvantages explained above.
In general we see a positive trend across all projects.
In the case of $\project_2$
and $\project_3$
our method performs decisively better than the reference by cycle $10$. For
$\project_1$
and $\project_4$
we meet or surpass the references in later cycles. 

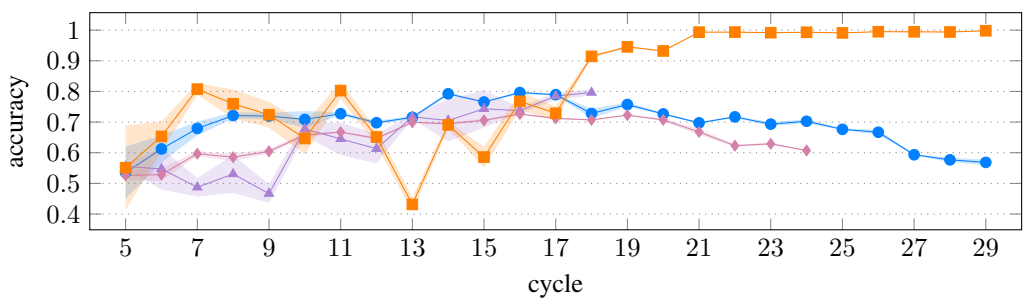
\begin{figure}
    \centering
    \begin{tikzpicture}
        \begin{axis}[
            xtick={5,7,...,29},
            ytick={0,0.1,...,1},
            xlabel=cycle,
            ylabel={accuracy},
            legend style={draw=none},
            width=\textwidth,
            height=.32\textwidth,
            ymajorgrids=true,
            grid style={dotted,gray},
            legend pos=south east,
            xmin=4,
            xmax=30,
            legend style={font=\tiny},
        ]
            \addplot[
                col3,
                mark=*,
                error bars/.cd, 
                y fixed,
                y dir=both, 
                y explicit,
            ] table [x index=0, y index=1] {data/project_1.dat};
            \errorband[col3, opacity=0.2]{data/project_1.dat}{cycle}{accuracy_mean}{accuracy_std};

            \addplot[
                col4,
                mark=triangle*,
                error bars/.cd, 
                y fixed,
                y dir=both, 
                y explicit,
            ] table [x index=0, y index=1] {data/project_2.dat};
            \errorband[col4, opacity=0.2]{data/project_2.dat}{cycle}{accuracy_mean}{accuracy_std};

            \addplot[
                col5,
                mark=diamond*,
                error bars/.cd, 
                y fixed,
                y dir=both, 
                y explicit,
            ] table [x index=0, y index=1] {data/project_3.dat};
            \errorband[col5, opacity=0.2]{data/project_3.dat}{cycle}{accuracy_mean}{accuracy_std}

            \addplot[
                col6,
                mark=square*,
                error bars/.cd, 
                y fixed,
                y dir=both, 
                y explicit,
            ] table [x index=0, y index=1] {data/project_4.dat};
            \errorband[col6, opacity=0.2]{data/project_4.dat}{cycle}{accuracy_mean}{accuracy_std}
            
        \end{axis}
    \end{tikzpicture}
    \caption{Ranking accuracy of our learned rewards. The dip in accuracy towards the end for some projects is explained by the increase in variance from smaller test datasets in later cycles.}
    \label{figure:accuracy}
\end{figure}

\section{Discussion}
\label{sec:discussion}
In this paper we have proposed a novel approach for automated reward configuration for drug discovery that relies solely on experimental data. We demonstrated the effectiveness of our approach when applied to historical data from real drug discovery projects, as well as in a simulated environment built on synthetic objectives taken from the GuacaMol benchmark.

The experiments on historical project data show that our method is able to learn functions that match or outperform the predictive power of human-configured functions when given access to the components from those functions, despite having access to inferior predictive models. In future work we will investigate the performance of our method when it is given access to all feasible components. This will allow us to determine if the constraint on available components applied in our experiments hinders or contributes towards the effectiveness of the approach. Applying our method to simulations of DMTA cycles showed that our method can be used to guide a generative molecular design tool towards molecules with scores comparable to those achieved in the
GuacaMol benchmark within $10$ iterations, despite seeding our runs with a randomly initialised population.

The contribution presented in this paper constitutes a significant step towards the automation of reward function configuration for generative molecular design, serving as a baseline for future research.

\bibliographystyle{plain} 
\bibliography{refs} 

\end{document}